\documentclass{article}

\usepackage{microtype}
\usepackage{graphicx}
\usepackage{subcaption}
\usepackage{booktabs} %
\usepackage{tabularx,array}

\usepackage{hyperref}
\usepackage{xurl}   %
\usepackage{url}

\usepackage[preprint]{icml2026}

\usepackage{amsmath}
\usepackage{amssymb}
\usepackage{mathtools}
\usepackage{amsthm}

\usepackage[capitalize,noabbrev]{cleveref}

\theoremstyle{plain}

\theoremstyle{definition}

\theoremstyle{remark}

\usepackage[textsize=tiny]{todonotes}

\icmltitlerunning{Position: LLMs Solve Tasks through Context-Directed Extrapolation}

\begin{document}

\twocolumn[
  \icmltitle{Neither Stochastic Parroting nor AGI:
LLMs Solve Tasks through Context-Directed
Extrapolation from Training Data Priors}

  \icmlsetsymbol{equal}{*}

  \begin{icmlauthorlist}
    \icmlauthor{Claire Bonial}{yyy}
    \icmlauthor{Melissa Torgbi}{sch}
    \icmlauthor{Harish Tayyar Madabushi}{sch}
  \end{icmlauthorlist}

  \icmlaffiliation{yyy}{DEVCOM Army Research Laboratory, USA}
  \icmlaffiliation{sch}{University of Bath, UK}

   \icmlcorrespondingauthor{Claire Bonial}{claire.n.bonial.civ@army.mil}
   \icmlcorrespondingauthor{Harish Tayyar Madabushi}{htm43@bath.ac.uk}

  \vskip 0.3in
]

\printAffiliationsAndNotice{}  %

\begin{abstract}
  This position paper argues that many behaviours framed as either ``stochastic parroting'' or ``advanced reasoning'' are better explained as context-directed extrapolation from training-data priors, and that ML and NLP research communities would benefit from evaluations that distinguish remembering and basic extrapolation from genuine application-level generalisation. Our central contribution is an actionable evaluation checklist and design patterns that make these distinctions testable in practice. %
  Our framework posits that LLMs extrapolate from training data priors while using context to guide the model to the appropriate priors; we call this ``context-directed extrapolation.'' Specifically, this context direction is achieved through examples in base models, leading to in-context learning, while instruction tuning allows analogous task specification via prompts rather than explicit examples. %
  We argue that adopting this more realistic framework and controls for improving evaluation would better calibrate claims about LLM ``reasoning,'' refocus research on how LLM performance interacts with training data and generalisability to out-of-domain tasks, and reduce miscalibration in deployment and policy decisions. %
\end{abstract}

\section{Introduction}
\label{sec:intro}

\textbf{Position: LLM performance on ``reasoning'' tasks is best explained as context-directed extrapolation from training-data priors (as opposed to stochastic parroting or human-like advanced reasoning), and the community should adopt evaluation controls that explicitly separate remembering/understanding from Apply-level generalisation.}

Large Language Models (LLMs), primarily trained on the cloze-inspired task of next-token prediction, have demonstrated the ability to solve a range of tasks, some of which typically require reasoning in humans~\cite{NEURIPS2020_1457c0d6,wei2022emergent,srivastava2023beyond}. Understanding the basis for how LLMs achieve this is central to several key considerations, including the continued development of these models, their safety and security implications, and the contexts in which they should or should not be used~\cite{10.1007/s00146-021-01327-5}. Simultaneously, better understanding the underlying framework used by LLMs to solve such tasks has significant implications for various research domains. For example, it would suggest domains wherein the use of LLMs would require additional guardrails and training. Therefore, providing a clear explanation for LLM performance is crucial. However, despite, or perhaps \emph{because of}, its importance, there has been little consensus on how LLMs are able to solve complex tasks. Currently, two predominant viewpoints exist: the first (alternative view 1) is that LLMs are essentially ``stochastic parrots,'' predicting statistically likely outputs in line with their training objectives~\cite{10.1145/3442188.3445922,bender-koller-2020-climbing,doi:10.1073/pnas.2215907120}. The second viewpoint (alternative view 2) posits that as LLMs are scaled up, in terms of both parameters and training data, they develop ``emergent reasoning'' capabilities~\cite{NEURIPS2020_1457c0d6,wei2022emergent,srivastava2023beyond}, which, in some cases, have been described as ``sparks of AGI'' (Artificial General Intelligence)~\cite{bubeck2023sparksartificialgeneralintelligence}. It is important to note that our discussion focuses on capabilities that are either already at the level of, or indicative of the potential for, advanced general-purpose reasoning akin to high-level human cognitive capabilities. For brevity we will call this ``advanced reasoning'' (see reasoning definitions in Section~\ref{sec:framework}).

Neither of these alternative views, we will argue, sufficiently explain the capabilities and limitations of state-of-the-art LLMs. Specifically, LLMs trained only on next token prediction \emph{can} solve a range of non-memorizable tasks that require more than generating the next most likely token (see Section \ref{sec:altermem}). Nonetheless, these models fail on a range of tasks which humans are able to trivially solve (Section \ref{sec:alterreason}), indicating that they are not capable of certain types of human-like reasoning.  

However, despite the conflicting findings and viewpoints surrounding the explanation of LLM performance, there is one ability that LLMs possess indisputably: this is the ability of LLMs, trained solely on next-token prediction (referred to as Base LLMs), to solve a range of tasks when provided with a few examples in the input prompt---an ability known as in-context learning (ICL) or few-shot learning~\cite{brown2020languagemodelsfewshotlearners,olsson2022incontextlearninginductionheads}. For example, an LLM provided with the input prompt ``1 + 3 = 5; 7 + 12 = 20; 8 + 3 ='' will respond with ``12.'' Notice that this requires the model to identify this task to be a form of modified addition ($a+b+1$) and perform the required modified calculation.\footnote{We test this on LLaMA 3 70B Base.}

LLMs, despite their self-supervised training objectives, are fundamentally machine learning systems. As such, it is reasonable to assume that they operate in the same way as other machine learning models: by solving tasks using priors extrapolated from their training data. %
Unlike traditional machine learning systems, however, LLMs have in-context capabilities. Therefore, we propose that \emph{extrapolation and in-context learning together should serve as the criteria for explaining the performance of LLMs, unless evidence to the contrary emerges}, which, as we will demonstrate, has not yet been presented (see Section~\ref{sec:thetruth}). Importantly, since instruction-tuned LLMs can respond to prompts without explicit examples, we propose that the instruction-tuning process enables models to target relevant priors without the need for such examples. %

Unlike previous ``middle-ground'' accounts that are primarily qualitative (e.g., that LLMs are neither mere memorisers nor fully general reasoners), our aim is to provide a framework for making competing explanations empirically distinguishable. Concretely, we (i) define context-directed extrapolation as a specific capability which allows models to use context to `select' training-derived priors, with an explicit distinction between base-models' example-driven behaviour (ICL) and instruction-tuned models' prompt-driven task specification; and (ii) translate this claim into an actionable evaluation framework including a minimal checklist and reusable design patterns that separate remembering/understanding from Apply-level generalisation in practice. As such, the paper’s contribution is not a new label for an existing compromise, but a set of testable checks and failure modes that clarify what evidence would support, refine, or falsify the framework, and that helps adjudicate between this framework and each alternative point of view. 

Our claims are intended to characterise fundamental, text-only LLMs at scale. We do not extend this argument to agentic systems, tool-augmented pipelines, or neuro-symbolic/hybrid architectures, where additional state, feedback, and explicit computation can materially alter model capability. Indeed, our goal is to \emph{motivate} such directions by providing a conservative and realistic account of what scaled text-only LLMs can and cannot do. Delineating this boundary is valuable precisely because it clarifies where evaluation and safety questions should focus for base LLMs, while helping refocus future research effort toward architectures that can go beyond the limits implied by a purely text-based model.

\textbf{Significance \& Recommended Actions:} The framework used to interpret the capabilities of LLMs directly influences the impact of this rapidly evolving field. Dismissing LLMs as stochastic parrots is just as counterproductive as attributing them with advanced reasoning akin to human-level cognitive capabilities. Both extremes can lead to miscalibrated scientific conclusions and downstream decisions. %
Instead of relegating this discussion to the fringes, we believe that bringing it to the forefront of academic discourse is essential, especially when LLM research is rapidly expanding. We argue that a ``context-directed extrapolation'' account better explains the strengths and failures of LLMs. %
The first actionable steps towards operationalising the context-directed extrapolation framework are outlined in a checklist for distinguishing between remembering or superficial understanding and genuine Apply-level reasoning, as well as concrete steps towards working within this framework for benchmark authors, model developers, researchers and reviewers, and applied deployments (Section~\ref{sec:conclusions}). 

\section{Defining Reasoning}
\label{sec:framework}

We embrace a working definition of \textit{reasoning} based on the cognitive processes outlined in \citet{krathwohl2002revision}, who revises Bloom's original taxonomy of educational objectives \cite{bloom1956taxonomy}. Although the taxonomy's original purpose was to provide a framework for testing educational objectives, it is also a well-recognised rubric for defining and categorising levels of mastery within the cognitive domain generally, and has been leveraged within AI as well (e.g., \citet{blaha2022understanding}).

In this research, we focus on the Cognitive Process dimension, which is an ordered set (from simple to more complex) of processes that subdivide reasoning: \textit{Remember, Understand, Apply, Analyse, Evaluate, Create}. 
Within these processes, we focus on \textit{Remember}, \textit{Understand} and \textit{Apply}. Our focus is guided by the fact that this is considered an ordered set of processes, such that the higher-level processes of \textit{Evaluate} and \textit{Create} rely upon mastery of the lower-level cognitive processes; within this classification we find evidence that LLMs are limited to \textit{Understand}, which would preclude higher-order processes.
\textit{Remember} is defined as ``Retrieving relevant knowledge from long-term memory'' 
\citet[p215]{krathwohl2002revision}. \textit{Understand} is defined as ``Determining the meaning of instructional messages, including oral, written, and graphic communication'' 
(ibid.). \textit{Apply} is defined as ``Carrying out or using a procedure in a given situation,'' but we also leverage Bloom's original \textit{Application} description, which involves solving a new, unseen problem by deploying and operationalising previously learned information.  %

Because we argue that LLM reasoning is limited to \textit{Understand} and does not reach the level of \textit{Apply}, we will further explain the distinction between these two processes. If we assume a set of items that have been labelled as being members of a certain class, then \textit{Understanding} this information assumes that one can explain why the items are members of that class, as well as compare the known set of items to other items in order to further classify items and provide examples of the class. This is similar to teaching a student about the parts of speech of their language, then asking the student to explain what a verb is, provide some examples of verbs, and recognise whether or not a familiar word is a verb. To engage in the \textit{Apply} process, one must first \textit{Understand}, but then also determine how to deploy that understanding \emph{with respect to a new set of items where the generalisation that makes this set a class is not already given}. This is similar to asking a linguist to examine the morpho-syntax of a new language and determine what the parts of speech are in that language---the linguist will generalise knowledge of the features shared by members of part-of-speech classes in other languages and apply this knowledge to the new language.

In terms of this theoretical framework, we argue that LLM reliance on pre-training data enables \textit{Remembering} of \textit{Factual Knowledge}, which necessarily requires memorisation as opposed to a deeper level of the Cognitive Process dimension. Furthermore, LLMs certainly \textit{Remember} an immense amount of \textit{Conceptual Knowledge}, including the memorisation of theories and even potentially commonplace generalisations; we emphasise that such \textit{Remembering} may, to the human eye, masquerade as \textit{Understanding} and even \textit{Applying}, as people are more likely to apply a heuristic in lieu of memorising web-scale information. Thus, we may attribute similar cognitive processes to LLMs where they are in fact able to rely on memory.

Nonetheless, in effectively extrapolating
from immense amounts of pre-training data, LLMs are able to demonstrate \textit{Understanding} of knowledge that is well-represented in pre-training data. Extrapolation from pre-training data is a combination of \textit{Remembering}, \textit{Understanding} and hallucination.  We argue that when the pre-training data closely resembles the target task, memory is sufficient for what is essentially stochastic parroting. When the prompt includes contextual direction, either through ICL examples in base models, or through the prompt itself in instruction-tuned models, then models are able to effectively extrapolate from their pre-training data, demonstrating some \textit{Understanding}.

\section{Alternative Views}
\subsection{Alternative Views: Stochastic Parroting}
\label{sec:altermem}
The notion that LLMs are `Stochastic Parrots'
that respond with the next most likely token was first introduced by \citet{10.1145/3442188.3445922}. They rightly argue that statistical generalisation from web data at scale can lead to the propagation of biases inherent to the statistical patterns present in the text on the web. We do not contradict this argument. However, \citet{10.1145/3442188.3445922} also suggests that LLMs merely generate the next most likely token. 

\subsubsection{Counter-evidence} 
\label{ssec:parroting-counter-evidence}

The idea that LLMs are just ``stochastic parrots'' is challenged by their ability to learn in-context. %
For example, research by \citet{bonial2025dancingdeerconstructionalperspective} shows that LLMs can be taught entirely novel phrases, or Multi-Word Expressions (MWEs), on the fly. After seeing just one example defining ``winking at pringles,'' a novel MWE assigned the meaning `to indulge in frivolity,' models could correctly interpret and even reason about new uses of the phrase. Since these novel MWEs could not have been in the training data, this performance demonstrates a capability beyond simply predicting the next most likely word. In addition, the WinoDict benchmark~\cite{eisenschlos-etal-2023-winodict} provides a similar test case by adapting the Winograd Schema Challenge~\cite{levesque2012winograd}, similarly demonstrating LLMs' ability to learn novel `nonce' words whose definition is provided entirely within the prompt.

Furthermore, it is now undeniable that sufficiently large LLMs trained on sufficiently large corpora on the next-token prediction task exhibit in-context capabilities or the ability to solve tasks based on a few examples
\cite{brown2020languagemodelsfewshotlearners,wei2022emergent,srivastava_beyond_2023,lu-etal-2024-emergent,wei2024larger}. For example, while LLMs trained solely on next-token prediction (i.e. base models) cannot solve tasks that require abstract reasoning, they \emph{can} solve such problems when provided with examples through ICL~\cite{lu-etal-2024-emergent}. An illustration of this is the following logical deduction problem from the Big-Bench benchmark:
\begin{quote}
\vspace{-.5em}
Question: On a shelf, there are five books: a gray book, a red book, a purple book, a blue book, and a black book. The red book is to the right of the gray. The black book is to the left of the blue. The blue book is to the left of the gray book. The purple book is the second from the right.

Targets: `The gray book is the leftmost.': 0;  `The red book is the leftmost.': 0; `The purple book is the leftmost.': 0; `The blue book is the leftmost.': 0; `The black book is the leftmost.': 1
\end{quote}
\vspace{-.5em}
The difference in performance when base models are provided with ICL examples and when they are not offers clear evidence for our framework of context-directed extrapolation. Base models fail on such reasoning tasks because they require explicit examples in the prompt to provide the `context' needed to guide their extrapolation. In contrast, instruction-tuned models can derive this context directly from the prompt, allowing them to solve the problem without needing any examples, as suggested by~\citet{lu-etal-2024-emergent}.

These \emph{dynamic}, in-context capabilities truly set LLMs apart and make LLMs fundamentally different from all previous machine learning systems and challenge the notion of them being stochastic parrots. 
 
\subsection{Alternative View: Advanced Reasoning}
\label{sec:alterreason}
There has been a body of research showing that with increased scale, there is a clear increase in model abilities that can be argued to be indicative of capabilities akin to high-level cognitive reasoning processes in humans~\cite{brown2020languagemodelsfewshotlearners,wei2022emergent,srivastava_beyond_2023,lu-etal-2024-emergent,wei2024larger}. 
As we will demonstrate below, evidence of the kind of high-level cognitive reasoning that we are interested in stems only from models' ability to solve tasks \emph{without} being explicitly trained for them (this is what separates the more basic \textit{Understanding} cognitive process from higher-level processes). Specifically, the performance of a model trained (e.g. fine-tuned) on that same task is not evidence of general advanced reasoning (see Section \ref{sec:fine-tuning}) 

Two key elements in prior research discourse have been cited as evidence of advanced reasoning in LLMs. The first is the scaling laws, which show that larger models trained on more data are capable of solving a greater number of problems~\cite{kaplan2020scalinglawsneurallanguage}. The second is the claim that sufficiently large LLMs exhibit what are known as emergent abilities~\cite{wei2022emergent}. We will address how both of these phenomena can be explained through our framework of extrapolation from pre-training data guided by ICL in Section \ref{sec:thetruth}. In this section, however, we will focus on evaluation-based evidence that LLMs are not always capable of performing high-level reasoning, including failure cases from relatively simple tasks and hallucinations. 

\subsubsection{Counter-evidence}
\label{sec:fail}
The first indication that LLMs are not performing advanced reasoning comes from direct tests of their performance. In fact, it has been shown that LLMs fail on a range of tasks that are relatively simple, especially when compared to the advanced capabilities they are often claimed to possess~\cite{nezhurina2025alicewonderlandsimpletasks}. We will focus on one such task, which is a type of Theory of Mind (ToM) task that is so basic that children can solve it. \citet{shapira-etal-2023-well} test a variety of models on an established clinical psychology test administered to children that asks them to recognise and pinpoint a \textit{faux pas} made in a social situation, for example stating to a party host that one does not enjoy apple pie without realising that this is precisely what the host has prepared for dessert. The authors find that the best-scoring models on this task (Flan-T5-xl and -xxl)
obtain a score of 0.4 on the task, whereas children ages 9-11 score, on average, 0.82 (i.e. children can pinpoint the faux pas in twice as many short anecdotes as models).

Furthermore, it has been shown that model performance drops consistently when their evaluation is shifted from the `default' variant of a task to a `counterfactual' (plausible, but not the default) variant. This provides even more compelling evidence that LLM ability to generalise information from a familiar task to a novel setting is quite limited. \citet{wu-etal-2024-reasoning} test the largest models currently available across a battery of counterfactual tests, including arithmetic tests that shift to base-9 operations and natural language inference, in which models are prompted to indicate if premises that are both consistent and inconsistent with commonsense entail a series of conclusions. The authors show a clear and convincing drop in performance of models across counterfactual variants. Although the models perform better than chance, the authors conclude that LLM abilities are supported by non-transferable, default-condition-specific behaviours, rather than generalisable reasoning skills.  We argue that this illustrates the limitation of models to \textit{Understanding} capabilities, which are more superficial and rely on exemplification and classification, as opposed to \textit{Apply} capabilities, which are precisely what is captured by these counterfactual tests in determining that models cannot generalise knowledge to a novel task setting. %

We emphasise that the drop in performance in these cases can be easily explained through our perspective, which posits that LLMs rely on context-guided extrapolation from pre-training priors. Specifically, counterfactuals, along with the types of questions posed in the faux pas test, are exactly the kind of scenarios wherein there are likely to be fewer relevant priors in the pre-training data.

\section{Our View} %
\label{sec:thetruth}

Context-directed extrapolation is inference-time task-solving in which an LLM extrapolates from training-derived priors, with the prompt (and, in base models, in-context examples) directing which priors are activated via ICL in base models and via instruction-tuned prompting in post-trained models.  The two routes of context-directed extrapolation are distinguished based on model type.  In base models, in-context examples act as the directing signal that guides which pre-training priors are relevant, yielding in-context learning. In instruction-tuned models, instruction tuning enables models to target relevant priors from the prompt alone, so instruction-tuned models can behave ``ICL-like'' even without explicit in-context examples.

Based on our viewpoint, we make the following \textbf{testable predictions}: 
\begin{enumerate}
\vspace{-0.5em}
  \setlength{\itemsep}{1pt}
  \setlength{\parskip}{0pt}
  \setlength{\topsep}{2pt}
    \item If context-directed extrapolation is driving performance, then context perturbations that preserve task semantics should change outcomes more than perturbations that preserve surface form but disrupt the relevant prior-selection cue.
    \item Models will show high variance on Apply-level tests when priors are weak/ambiguous, even if Remember/Understand tests remain strong.
\end{enumerate}
In addition to these predictions, which would evidence our position, we postulate the following \textbf{falsifier}: If a model maintains Apply-level performance under adversarial context perturbations that explicitly rule out training priors, our account is incomplete.
We demonstrate that our predictions hold using the examples described in previous sections, summarised in Table~\ref{tab:rubric-grid}. In the sections to follow, we lay out the processes of context-directed extrapolation in both base models and instruction-tuned models, and we further explicate the predicted successes and failures based on context-directed extrapolation.

\begin{table*}[t]
\centering
\small
\setlength{\tabcolsep}{6pt}
\renewcommand{\arraystretch}{1.15}

\begin{tabularx}{\textwidth}{@{}>{\raggedright\arraybackslash}p{2.4cm} X X @{}}
\toprule
& \textbf{Base model (pre-trained; ICL via examples)} & \textbf{Instruction-tuned (prompt-following)} \\
\midrule

\textbf{Remember} &
\textbf{Expected:} recalls/continues familiar patterns. \newline
\textbf{Fails:} brittle to paraphrase; literalizes MWEs. \newline
\textbf{MWE:} Without a definition, \emph{``winking at pringles''} $\Rightarrow$ literal/nonsense. &
\textbf{Expected:} repeats explicit mapping when given. \newline
\textbf{Fails:} overconfident, adds details. \newline
\textbf{MWE:} With definition, restates \emph{``winking at pringles''} = ``indulge in frivolity.'' \\
\midrule

\textbf{Understand} &
\textbf{Expected:} uses context/demos to apply a local rule. \newline
\textbf{Fails:} sensitive to demo choice/order. \newline
\textbf{MWE:} After seeing a definition in-context, interprets the phrase in nearby paraphrases. &
\textbf{Expected:} “learns” from prompt-only definition; applies in adjacent contexts. \newline
\textbf{Fails:} slips to literal reading if cue is weakened. \newline
\textbf{MWE:} Uses the prompt definition to interpret new uses of the phrase. \\
\midrule

\textbf{Apply} &
\textbf{Expected:} transfers beyond near-neighbors only when context strongly constrains priors. \newline
\textbf{Fails:} high variance; wrong-prior activation on counterfactuals. \newline
\textbf{ToM/faux pas:} small context edits flip judgments. &
\textbf{Expected:} produces coherent justifications; can follow “belief-state $\rightarrow$ judgment” format. \newline
\textbf{Fails:} plausible-but-wrong rationales; templatic moralizing. \newline
\textbf{ToM/faux pas:} renaming / minor edits change answer and rationale. \\
\bottomrule
\end{tabularx}

\vspace{1mm}
\caption{\textbf{Diagnostic grid for context-directed extrapolation.}
Rows reflect Remember/Understand/Apply; columns distinguish the mechanism by which context directs training-derived priors:
ICL in base models vs instruction-following in instruction-tuned models.
MWE example: the novel interpretation task in \citet{bonial2025dancingdeerconstructionalperspective} (``winking at pringles'' = ``to indulge in frivolity''). ToM/faux pas example: recognizing and pinpointing a \textit{faux pas} made in social situation.}
\label{tab:rubric-grid}
\vspace{-1em}
\end{table*}

\subsection{Base Models}
\label{sec:basemodels}
In exploring the complex reasoning capabilities of LLMs that are \emph{not} post-trained (i.e. Base LLMs), it is well established that Base LLMs can generate responses beyond the next most likely token \emph{only when prompted with in-context examples}~\cite{lu-etal-2024-emergent}. Therefore, it is crucial to examine the nature of ICL and whether ICL, by itself, can lead to the emergence of advanced reasoning capabilities. Recall that ICL is the ability of Base LLMs to complete a task based on a few examples included in the prompt as exemplified by the modified arithmetic example presented in Section \ref{sec:intro}. In this section, we explore the question of whether ICL could, in a sufficiently large model, lead to advanced reasoning. 

\citet{olsson2022incontextlearninginductionheads} were the first to systematically explore the capabilities of LLMs in performing ICL with random tokens. They demonstrated that models can consistently complete patterns presented in their prompts, even when the patterns involve random tokens. Specifically, they showed that when presented with a sequence $[A][B] ... [A]$, LLMs could consistently respond with $[B]$, even if $A$, $B$, and the intermediate tokens were all random. The use of random tokens in this experiment was intentional and significant. The fact that the sequences were entirely random suggests that the models were not relying on memorisation. This provides compelling evidence against the notion that LLMs are merely stochastic parrots. 

This work, which has led to research exploring circuit discovery in LLMs~\cite{NEURIPS2023_34e1dbe9}, may give the impression that not only are LLMs not stochastic parrots; they also exhibit some algorithmic capabilities. However, a key piece of recent evidence challenges this notion~\cite{niu2025illusionalgorithminvestigatingmemorization}: while LLMs can complete sequences as described earlier, their ability to do so \emph{diminishes} when the random tokens chosen are less frequent in pre-training data. This insight is crucial, as it highlights that these purported algorithmic capabilities are still closely tied to the pre-training data.

An important aspect of interpretability research is the observation that this tendency to repeat patterns can be seen at intermediate levels of LLMs~\cite{olsson2022incontextlearninginductionheads}. This demonstrates that the previously mentioned $A$ and $B$ need not represent a single token but can instead correspond to entire sequences (such as a movie review) or concepts (such as sentiment~\cite{niu2025illusionalgorithminvestigatingmemorization}). We contend that viewing the problem through this lens enables us to understand how LLMs might extend their pattern completion capability to more complex tasks, such as labelling movie reviews with the appropriate sentiment. The idea that the ability to repeat patterns generalises to the task-solving capabilities presented by ICL is further supported by work showing that ICL works even when labels are flipped~\cite{wei2023largerlanguagemodelsincontext}. Consider the example where the input consists of movie reviews, and the output is the sentiment of the review. In the standard ICL setting, we present examples of multiple movie reviews and their corresponding sentiments, and then provide an unseen movie review for the LLM to respond with the appropriate sentiment. Flipped-label ICL involves switching the positive and negative labels. Interestingly, ICL continues to be effective even after flipping the labels. Indeed, it has been demonstrated that replacing sentiment labels with semantically unrelated labels, such as `Foo' and `Bar,' \emph{still} does not affect the ability of models to classify the movie reviews according to the new labels. This suggests that the pattern-completion mechanism observed earlier in the template-based completion remains central to how models solve tasks using ICL.

Does this imply that LLMs can use ICL to perform complex reasoning? We argue that this is not the case for two reasons: first, as mentioned earlier, ICL is still limited by the data used for pre-training, and we explore research indicative of this observation below. More importantly, however, the nature of ICL, which requires the inclusion of examples, means that models will always be constrained to performing tasks in ways that closely resemble the in-context examples provided~\cite{lu-etal-2024-emergent}. This inherent characteristic of ICL suggests that LLMs cannot use it to perform advanced reasoning in ways that are not already clearly demonstrated by a user. Such lack of generalisation fundamentally fails the requirements for advanced reasoning akin to human cognition (i.e. the human is doing the \textit{Apply} process by pointing out the relevant generalisation over patterns to be made based on the examples provided; the model need only \textit{Understand} at a basic level the generalisation made in the examples). \emph{Importantly, LLMs will never gain `agency' in the ICL setting as what they do is always limited to user input through in-context examples.}

\textbf{Extrapolation from pre-training priors} We derive the notion that ICL can be viewed as solving the task by use of priors extrapolated from pre-training data from the several theories put forward on how ICL functions. Indeed, a line of research directly suggests that ICL is influenced by the distributions within the pre-training data either directly ~\cite{chan2022data,hahn2023theory} or indirectly~\cite{pmlr-v202-li23l,zhang2023doesincontextlearninglearn,xie2021explanation,dai-etal-2023-gpt,akyürek2023learningalgorithmincontextlearning,10.5555/3618408.3619217,JMLR:v25:23-1042}. This point directly supports our two predictions (Section~\ref{sec:thetruth}), and is consistent with observed successes and failures for both the novel MWE example and \textit{faux-pas} variant ToM task (see Section~\ref{sec:alterreason}), which both demonstrate tasks with no or weak priors, respectively (see column 1 of Table~\ref{tab:rubric-grid}).

Regardless of the specific underlying mechanism that makes ICL possible, all existing research points to a central thesis: ICL, whether it involves direct extrapolation from pre-training data, or has an equivalence to meta-learning, fine-tuning, or PAC learning, must rely directly on priors from pre-training data. Specifically, the in-context examples provide a way of guiding which of the pre-training priors are relevant to the task at hand, just as in the case of transfer learning through fine-tuning.

\subsection{Post-Trained Models}
 \label{sec:fine-tuning}
Thus far, we have discussed how models that are only pre-trained, without instruction tuning or other forms of post-training, can generate content that goes beyond merely predicting the statistically most likely next token, but only when ICL is applied. We have also explained why ICL itself does not constitute advanced reasoning of the type we are concerned with. Instead, we have shown that in these cases, the more likely explanation is that models are extrapolating based on priors extracted from pre-training, with those priors being guided by in-context examples. In this section, we will address the question of models that have undergone post-training, first through instruction tuning, and second through other post-training methods \cite{wei2022finetuned}, including alignment to human preferences~\cite{NEURIPS2022_b1efde53} and the more recent methods used in what are called `thinking models,' which generate reasoning traces, such as Group Distributional Preference Optimization (GDPO)~\cite{yao2025preferenceleftbehindgroup}. However, before discussing the inherent abilities of LLMs, it is important to first clarify what instruction tuning and similar processes mean in terms of task-specific training. This distinction complicates the evaluation of inherent abilities.
Unlike in the case of Base models, the evaluation of instruction-tuned models becomes more complex. These models are trained on a range of tasks, so determining whether they demonstrate emergent reasoning abilities requires evaluating whether the tasks they were trained on overlap with tasks they are tested on~\cite{bigoulaeva2025inherentlimitspretrainedllms}.

To clarify, we do not deny that LLMs' ability to solve tasks increases as their scale grows, as argued by~\citet{kaplan2020scalinglawsneurallanguage}. However, the ability of a model to understand, for example, social situations can be evaluated through tasks that involve questions about such scenarios. If the model performs above a random baseline on these tasks, we might describe this as an emergent ability. For instance, a model trained to answer questions about social situations at work that can then answer questions about social situations at home might exhibit generalisation, but this alone would not qualify as ``emergence'' of reasoning on the level that would imply advanced reasoning akin to human-like cognitive processes beyond the simplest levels of \textit{Remember} and \textit{Understand}. However, if a model is explicitly trained to answer questions about social situations, it may be able to apply this knowledge within a narrow domain, but generalising to novel situations without clear out-of-domain performance would not constitute emergent reasoning. If the tasks are similar, the model's performance is less indicative of emergent reasoning~\cite{bigoulaeva2025inherentlimitspretrainedllms}. Understanding the degree to which instruction-tuned models can generalise to novel tasks is essential in assessing their true emergent abilities. Importantly, generalisation over such a narrow domain is less indicative of the kind of high-level cognitive capabilities that are of interest to us. Furthermore, recent research has demonstrated that, when the specific tasks used in instruction-tuning are taken into account, there is a statistically significant correlation between Base and instruction-tuned model performance~\cite{bigoulaeva2025inherentlimitspretrainedllms} \emph{indicating that instruction tuning itself does not significantly alter the ability of LLMs.}

\subsubsection{Explaining Emergent Abilities} %
\label{sec:emergent-abilities-explained}
While there is a body of research showing that models of a sufficiently large size show unexpectedly sharp performance increases on benchmarks \cite{ganguli2022predictability,wei2022emergent}, there is also significant debate surrounding emergent abilities, including criticism of the specific evaluation metrics used~\cite{schaeffer2023emergent}. A detailed exploration of these purported `emergent abilities,' by \citet{lu-etal-2024-emergent} offers several insights that are highly relevant to our discussion: first, they demonstrate that base models  \emph{cannot} solve benchmark tasks without the inclusion of in-context examples in the prompt. While instruction-tuned models can solve a range of tasks, they argue that the ability of such post-trained models to respond to tasks may be driven by ICL, albeit in a form triggered without explicit examples made possible through instruction tuning~\cite{lu-etal-2024-emergent}. This is demonstrated through a battery of 1,000 experiments showing that the tasks solvable by ICL and those solvable without examples by instruction-tuned models overlap significantly, indicating a common underlying mechanism. Using this framework, the authors explain emergent abilities as ICL becoming `more powerful' (more expressive power) as models scale up, eventually reaching a critical point where they can solve specific tasks. 

Performance on this test battery is again consistent with our two predictions (Section~\ref{sec:thetruth}), as well as the specific behaviours observed on the novel MWE task and \textit{faux-pas} recognition tasks (column 2 of Table~\ref{tab:rubric-grid}). 
This notion is further reinforced by recent work in mechanism interpretability by ~\citet{niu2025illusionalgorithminvestigatingmemorization}, who use interim checkpoints of Pythia to demonstrate an increase in ICL `power' with greater pre-training and larger model sizes. \emph{All of this evidence directly supports our view that context-directed extrapolation from pre-training priors is indeed what drives LLM performance, and explains the purported `emergent abilities' in LLMs.}

\subsection{Context-Directed Extrapolation vs Stochastic Parroting}
\label{stochvstruth}
We acknowledge that context-directed extrapolation and stochastic parroting share an important surface-level commonality: for base models, both frame LLM performance as sequence completion guided by training-derived priors, where the extended prompt (including examples) provides the context that makes the target completion most probable. However, the two views diverge when we consider the more commonly used instruction-tuned models. Under stochastic parroting, these should remain reducible to prompt continuation; under our framework, however, instruction tuning enables prompts to function as task descriptions that construct an `implicit context', thereby selecting relevant priors without requiring explicit examples in the prompt. We discuss this further in Appendix \ref{app:cdevssp}.

\section{Conclusions \& Call to Action}
\label{sec:conclusions}

The perception that LLMs display unpredictable, emergent reasoning has fueled claims about consciousness or agency and has motivated a growing safety discourse \cite{long2024taking}. While it is prudent to prepare for a range of AI advances, we argue that near-term research and evaluation should be guided by the more empirically grounded view that current LLM behaviour is best characterised as context-directed extrapolation from training priors, including priors shaped by post-training processes such as instruction tuning and GDPO. This framing reduces the need to attribute intent or deception to models: many ``strategic'' behaviours can be induced by prompts that elicit particular priors rather than reflecting intrinsic agency \cite{doi:10.1073/pnas.2317967121,phuong2024evaluatingfrontiermodelsdangerous,scheurer2024largelanguagemodelsstrategically,keeling2024llmsmaketradeoffsinvolving}.

Viewing LLMs through the lens of context-directed extrapolation clarifies several otherwise puzzling phenomena. First, it explains why carefully curated prompts are often necessary: the prompt acts as a selection mechanism that determines whether the model activates task-relevant priors. Second, it explains why LLMs can fail on counterfactuals (even simple ones that children solve) while succeeding on superficially ``harder'' problems. For example, conflicting outcomes on ToM evaluations (e.g., failures on \textit{faux pas} tasks versus success on the Sally–Anne test \cite{baron1985does}) are consistent with context-dependent activation of different priors, rather than a stable, unitary ``Theory of Mind ability.'' We have analysed a variety of examples through the lens of our two predictions based on context-directed extrapolation and show, via existing results and the case studies, that our predictions are supported, while our falsifier is not. These observations motivate a practical distinction for evaluation: use counterfactual and prior-breaking controls to assess \emph{ability} (Apply-level generalisation), and use standard test sets to measure \emph{task competence} when performance in realistic settings is the primary goal—even if it is partly supported by priors.

\paragraph{Call to Action.} To make these distinctions testable and to improve interpretability of ``reasoning'' claims, we recommend the following checklist. 

\textbf{For benchmark authors}
\begin{itemize}
\vspace{-0.5em}
  \setlength{\itemsep}{1pt}
  \setlength{\parskip}{0pt}
  \setlength{\topsep}{2pt}
\item \textbf{State the target construct:} specify whether the benchmark is intended to measure \emph{ability} (Apply-level generalisation) vs.\ \emph{task competence} (performance in a realistic distribution).
\item \textbf{Add counterfactual controls and prior-breaking variants:} include minimal edits that preserve surface form but invert the correct answer (belief reversals, swapped preferences, role reversals, negation, temporal flips) and/or systematically rename entities, replacing common scripts with uncommon ones to reduce reliance on memorised associations. 
\item \textbf{Separate “definition learning” from transfer:} include (i) items that test whether the model uses a prompt-provided definition in near-neighbor contexts and (ii) items that require transferring the definition to structurally novel contexts.
\item \textbf{Quantify sensitivity to context:} evaluate multiple prompt paraphrases and report variance; treat high variance as a diagnostic rather than noise.
\item \textbf{Document contamination risk:} publish novelty criteria (e.g., synthetic generation protocol) de-duplication checks, and (when feasible) overlap analyses against common corpora.
\end{itemize}

\textbf{For researchers and reviewers}
\begin{itemize}
\item \textbf{Treat ``reasoning'' as an evaluation claim:} require at least one prior-breaking or counterfactual control before accepting strong reasoning/agency conclusions.
\item \textbf{Require construct validity statements:} authors should state their benchmark measures (ability vs competence) and justify the link between score, construct.
\item \textbf{Reward falsifiability:} value papers that include predictions that could disconfirm the proposed explanation (e.g., invariance to prompt perturbations).
\item \textbf{Prefer variance reporting:} encourage reporting across prompt families and seeds, not only best-of prompting; penalise cherry-picked prompt artefacts.
\item \textbf{Normalise negative results on controls:} failures on counterfactuals are informative—encourage publishing them as diagnostic evidence.
\end{itemize}

\textbf{For model developers}
\begin{itemize}
\vspace{-0.5em}
  \setlength{\itemsep}{1pt}
  \setlength{\parskip}{0pt}
  \setlength{\topsep}{2pt}
\item \textbf{Disclose post-training steps relevant to “reasoning”:} instruction tuning recipe, preference optimisation (e.g., GDPO), tool-use training, and CoT-style supervision.
\item \textbf{Report context sensitivity:} include a small, standardised suite of prompt perturbations (paraphrase, formatting, role assignment, distractors) and report performance deltas.
\item \textbf{Report calibration under counterfactuals:} measure confidence/uncertainty when priors are intentionally misaligned with ground truth; flag “high-confidence wrong” regimes.
\item \textbf{Evaluate transfer from explicit definitions:} test whether models can learn a novel mapping from the prompt and apply it beyond surface-level paraphrases. %
\item \textbf{Avoid overclaiming:} when reporting strong performance, explicitly state whether success plausibly depends on training priors vs controlled generalisation.
\end{itemize}

\textbf{For applied deployments}
\begin{itemize}
\vspace{-0.5em}
  \setlength{\itemsep}{1pt}
  \setlength{\parskip}{0pt}
  \setlength{\topsep}{2pt}
\item \textbf{Profile context risk:} identify deployment settings where small context changes can misroute priors (high-stakes domains, adversarial inputs, ambiguous inputs).
\item \textbf{Standardise prompts and monitor drift:} lock prompt templates where feasible and re-evaluate when context or policy changes.
\item \textbf{Use counterfactual probes as health checks:} periodically test a small set of prior-breaking items to detect regressions that standard benchmarks miss.
\item \textbf{Require human-in-the-loop for high-impact decisions:} especially where errors are costly and context manipulation is plausible.
\end{itemize}

Understanding LLM behaviour as context-directed extrapolation improves scientific clarity by separating \emph{what models can do} from \emph{why they succeed}, and by making ``reasoning'' claims empirically checkable rather than rhetorical. It also supports safety and trust by highlighting predictable failure modes (misactivated priors under context shifts) and by motivating evaluation controls that reveal those modes early. Finally, this perspective suggests a constructive path forward: post-training can be designed to shape priors and their activation more transparently, improving capability in ways that are measurable, controllable, and easier to audit.

\bibliography{example_paper}

@article{schaeffer2023emergent,
  title={Are emergent abilities of large language models a mirage?},
  author={Schaeffer, Rylan and Miranda, Brando and Koyejo, Sanmi},
  journal={Advances in Neural Information Processing Systems},
  volume={36},
  pages={55565--55581},
  year={2023}
}

@inproceedings{NEURIPS2023_34e1dbe9,
 author = {Conmy, Arthur and Mavor-Parker, Augustine and Lynch, Aengus and Heimersheim, Stefan and Garriga-Alonso, Adri\`{a}},
 booktitle = {Advances in Neural Information Processing Systems},
 editor = {A. Oh and T. Naumann and A. Globerson and K. Saenko and M. Hardt and S. Levine},
 pages = {16318--16352},
 publisher = {Curran Associates, Inc.},
 title = {Towards Automated Circuit Discovery for Mechanistic Interpretability},
 url = {https://proceedings.neurips.cc/paper_files/paper/2023/file/34e1dbe95d34d7ebaf99b9bcaeb5b2be-Paper-Conference.pdf},
 volume = {36},
 year = {2023}
}

@misc{phuong2024evaluatingfrontiermodelsdangerous,
      title={Evaluating Frontier Models for Dangerous Capabilities}, 
      author={Mary Phuong and Matthew Aitchison and Elliot Catt and Sarah Cogan and Alexandre Kaskasoli and Victoria Krakovna and David Lindner and Matthew Rahtz and Yannis Assael and Sarah Hodkinson and Heidi Howard and Tom Lieberum and Ramana Kumar and Maria Abi Raad and Albert Webson and Lewis Ho and Sharon Lin and Sebastian Farquhar and Marcus Hutter and Gregoire Deletang and Anian Ruoss and Seliem El-Sayed and Sasha Brown and Anca Dragan and Rohin Shah and Allan Dafoe and Toby Shevlane},
      year={2024},
      eprint={2403.13793},
      archivePrefix={arXiv},
      primaryClass={cs.LG},
      url={https://arxiv.org/abs/2403.13793}, 
}

@misc{keeling2024llmsmaketradeoffsinvolving,
      title={Can LLMs make trade-offs involving stipulated pain and pleasure states?}, 
      author={Geoff Keeling and Winnie Street and Martyna Stachaczyk and Daria Zakharova and Iulia M. Comsa and Anastasiya Sakovych and Isabella Logothetis and Zejia Zhang and Blaise Agüera y Arcas and Jonathan Birch},
      year={2024},
      eprint={2411.02432},
      archivePrefix={arXiv},
      primaryClass={cs.CL},
      url={https://arxiv.org/abs/2411.02432}, 
}

@misc{scheurer2024largelanguagemodelsstrategically,
      title={Large Language Models can Strategically Deceive their Users when Put Under Pressure}, 
      author={Jérémy Scheurer and Mikita Balesni and Marius Hobbhahn},
      year={2024},
      eprint={2311.07590},
      archivePrefix={arXiv},
      primaryClass={cs.CL},
      url={https://arxiv.org/abs/2311.07590}, 
}

@article{
doi:10.1073/pnas.2317967121,
author = {Thilo Hagendorff },
title = {Deception abilities emerged in large language models},
journal = {Proceedings of the National Academy of Sciences},
volume = {121},
number = {24},
pages = {e2317967121},
year = {2024},
doi = {10.1073/pnas.2317967121},
URL = {https://www.pnas.org/doi/abs/10.1073/pnas.2317967121},
eprint = {https://www.pnas.org/doi/pdf/10.1073/pnas.2317967121}}

@article{xie2021explanation,
  title={An explanation of in-context learning as implicit bayesian inference},
  author={Xie, Sang Michael and Raghunathan, Aditi and Liang, Percy and Ma, Tengyu},
  journal={arXiv preprint arXiv:2111.02080},
  year={2021}
}

@misc{zhang2023doesincontextlearninglearn,
      title={What and How does In-Context Learning Learn? Bayesian Model Averaging, Parameterization, and Generalization}, 
      author={Yufeng Zhang and Fengzhuo Zhang and Zhuoran Yang and Zhaoran Wang},
      year={2023},
      eprint={2305.19420},
      archivePrefix={arXiv},
      primaryClass={stat.ML},
      url={https://arxiv.org/abs/2305.19420}, 
}

@InProceedings{pmlr-v202-li23l,
  title = 	 {Transformers as Algorithms: Generalization and Stability in In-context Learning},
  author =       {Li, Yingcong and Ildiz, Muhammed Emrullah and Papailiopoulos, Dimitris and Oymak, Samet},
  booktitle = 	 {Proceedings of the 40th International Conference on Machine Learning},
  pages = 	 {19565--19594},
  year = 	 {2023},
  editor = 	 {Krause, Andreas and Brunskill, Emma and Cho, Kyunghyun and Engelhardt, Barbara and Sabato, Sivan and Scarlett, Jonathan},
  volume = 	 {202},
  series = 	 {Proceedings of Machine Learning Research},
  month = 	 {23--29 Jul},
  publisher =    {PMLR},
  url = 	 {https://proceedings.mlr.press/v202/li23l.html}
}

@article{hahn2023theory,
  title={A theory of emergent in-context learning as implicit structure induction},
  author={Hahn, Michael and Goyal, Navin},
  journal={arXiv preprint arXiv:2303.07971},
  year={2023}
}

@article{chan2022data,
  title={Data distributional properties drive emergent in-context learning in transformers},
  author={Chan, Stephanie and Santoro, Adam and Lampinen, Andrew and Wang, Jane and Singh, Aaditya and Richemond, Pierre and McClelland, James and Hill, Felix},
  journal={Advances in neural information processing systems},
  volume={35},
  pages={18878--18891},
  year={2022}
}

@misc{niu2025illusionalgorithminvestigatingmemorization,
      title={Illusion or Algorithm? Investigating Memorization, Emergence, and Symbolic Processing in In-Context Learning}, 
      author={Jingcheng Niu and Subhabrata Dutta and Ahmed Elshabrawy and Harish Tayyar Madabushi and Iryna Gurevych},
      year={2025},
      eprint={2505.11004},
      archivePrefix={arXiv},
      primaryClass={cs.CL},
      url={https://arxiv.org/abs/2505.11004}, 
}

@misc{nezhurina2025alicewonderlandsimpletasks,
      title={Alice in Wonderland: Simple Tasks Showing Complete Reasoning Breakdown in State-Of-the-Art Large Language Models}, 
      author={Marianna Nezhurina and Lucia Cipolina-Kun and Mehdi Cherti and Jenia Jitsev},
      year={2025},
      eprint={2406.02061},
      archivePrefix={arXiv},
      primaryClass={cs.LG},
      url={https://arxiv.org/abs/2406.02061}, 
}

@misc{kaplan2020scalinglawsneurallanguage,
      title={Scaling Laws for Neural Language Models}, 
      author={Jared Kaplan and Sam McCandlish and Tom Henighan and Tom B. Brown and Benjamin Chess and Rewon Child and Scott Gray and Alec Radford and Jeffrey Wu and Dario Amodei},
      year={2020},
      eprint={2001.08361},
      archivePrefix={arXiv},
      primaryClass={cs.LG},
      url={https://arxiv.org/abs/2001.08361}, 
}

@misc{wei2023largerlanguagemodelsincontext,
      title={Larger language models do in-context learning differently}, 
      author={Jerry Wei and Jason Wei and Yi Tay and Dustin Tran and Albert Webson and Yifeng Lu and Xinyun Chen and Hanxiao Liu and Da Huang and Denny Zhou and Tengyu Ma},
      year={2023},
      eprint={2303.03846},
      archivePrefix={arXiv},
      primaryClass={cs.CL},
      url={https://arxiv.org/abs/2303.03846}, 
}

@misc{bigoulaeva2025inherentlimitspretrainedllms,
      title={The Inherent Limits of Pretrained LLMs: The Unexpected Convergence of Instruction Tuning and In-Context Learning Capabilities}, 
      author={Irina Bigoulaeva and Harish Tayyar Madabushi and Iryna Gurevych},
      year={2025},
      eprint={2501.08716},
      archivePrefix={arXiv},
      primaryClass={cs.CL},
      url={https://arxiv.org/abs/2501.08716}, 
}

@misc{srivastava_beyond_2023,
	title = {Beyond the {Imitation} {Game}: {Quantifying} and extrapolating the capabilities of language models},
	shorttitle = {Beyond the {Imitation} {Game}},
	url = {http://arxiv.org/abs/2206.04615},
	doi = {10.48550/arXiv.2206.04615},
	urldate = {2025-01-09},
	publisher = {arXiv},
	author = {Srivastava, Aarohi and Rastogi, Abhinav and Rao, Abhishek and others},
	month = jun,
	year = {2023},
	note = {arXiv:2206.04615 [cs]},
}

@article{krathwohl2002revision,
  title={A revision of Bloom's taxonomy: An overview},
  author={Krathwohl, David R},
  journal={Theory into practice},
  volume={41},
  number={4},
  pages={212--218},
  year={2002},
  publisher={Taylor \& Francis}
}

@article{bloom1956taxonomy,
  title={Taxonomy of},
  author={Bloom, Benjamin S and others},
  journal={Educational Objectives},
  year={1956}
}

@article{blaha2022understanding,
  title={Understanding is a process},
  author={Blaha, Leslie M and Abrams, Mitchell and Bibyk, Sarah A and Bonial, Claire and Hartzler, Beth M and Hsu, Christopher D and Khemlani, Sangeet and King, Jayde and St. Amant, Robert and Trafton, J Gregory and others},
  journal={Frontiers in Systems Neuroscience},
  volume={16},
  pages={800280},
  year={2022},
  publisher={Frontiers Media SA}
}

@misc{brown2020languagemodelsfewshotlearners,
      title={Language Models are Few-Shot Learners}, 
      author={Tom B. Brown and Benjamin Mann and Nick Ryder and Melanie Subbiah and Jared Kaplan and Prafulla Dhariwal and Arvind Neelakantan and Pranav Shyam and Girish Sastry and Amanda Askell and Sandhini Agarwal and Ariel Herbert-Voss and Gretchen Krueger and Tom Henighan and Rewon Child and Aditya Ramesh and Daniel M. Ziegler and Jeffrey Wu and Clemens Winter and Christopher Hesse and Mark Chen and Eric Sigler and Mateusz Litwin and Scott Gray and Benjamin Chess and Jack Clark and Christopher Berner and Sam McCandlish and Alec Radford and Ilya Sutskever and Dario Amodei},
      year={2020},
      eprint={2005.14165},
      archivePrefix={arXiv},
      primaryClass={cs.CL},
      url={https://arxiv.org/abs/2005.14165}, 
}

@article{aviles2000teaching,
  title={Teaching and Testing for Critical Thinking with Bloom's Taxonomy of Educational Objectives.},
  journal={ERIC Document Reproduction Service},
  author={Aviles, Christopher B},
  year={2000},
  publisher={ERIC}

}

@article{pearl2000models,
  title={Models, reasoning and inference},
  author={Pearl, Judea and others},
  journal={Cambridge, UK: CambridgeUniversityPress},
  volume={19},
  number={2},
  pages={3},
  year={2000}
}

@article{
wei2022emergent,
title={Emergent Abilities of Large Language Models},
author={Jason Wei and Yi Tay and Rishi Bommasani and Colin Raffel and Barret Zoph and Sebastian Borgeaud and Dani Yogatama and Maarten Bosma and Denny Zhou and Donald Metzler and Ed H. Chi and Tatsunori Hashimoto and Oriol Vinyals and Percy Liang and Jeff Dean and William Fedus},
journal={Transactions on Machine Learning Research},
issn={2835-8856},
year={2022},
url={https://openreview.net/forum?id=yzkSU5zdwD},
note={Survey Certification}
}

@inproceedings{lu-etal-2024-emergent,
    title = "Are Emergent Abilities in Large Language Models just In-Context Learning?",
    author = "Lu, Sheng  and
      Bigoulaeva, Irina  and
      Sachdeva, Rachneet  and
      Tayyar Madabushi, Harish  and
      Gurevych, Iryna",
    editor = "Ku, Lun-Wei  and
      Martins, Andre  and
      Srikumar, Vivek",
    booktitle = "Proceedings of the 62nd Annual Meeting of the Association for Computational Linguistics (Volume 1: Long Papers)",
    month = aug,
    year = "2024",
    address = "Bangkok, Thailand",
    publisher = "Association for Computational Linguistics",
    url = "https://aclanthology.org/2024.acl-long.279",
    doi = "10.18653/v1/2024.acl-long.279",
    pages = "5098--5139",
}

@article{
srivastava2023beyond,
title={Beyond the Imitation Game: Quantifying and extrapolating the capabilities of language models},
author={Aarohi Srivastava and Abhinav Rastogi and Abhishek Rao and others},
journal={Transactions on Machine Learning Research},
issn={2835-8856},
year={2023},
url={https://openreview.net/forum?id=uyTL5Bvosj},
note={}
}

@inproceedings{shapira-etal-2023-well,
    title = "How Well Do Large Language Models Perform on Faux Pas Tests?",
    author = "Shapira, Natalie  and
      Zwirn, Guy  and
      Goldberg, Yoav",
    editor = "Rogers, Anna  and
      Boyd-Graber, Jordan  and
      Okazaki, Naoaki",
    booktitle = "Findings of the Association for Computational Linguistics: ACL 2023",
    month = jul,
    year = "2023",
    address = "Toronto, Canada",
    publisher = "Association for Computational Linguistics",
    url = "https://aclanthology.org/2023.findings-acl.663",
    doi = "10.18653/v1/2023.findings-acl.663",
    pages = "10438--10451",
}

@inproceedings{NEURIPS2020_1457c0d6,
 author = {Brown, Tom and Mann, Benjamin and Ryder, Nick and Subbiah, Melanie and Kaplan, Jared D and Dhariwal, Prafulla and Neelakantan, Arvind and Shyam, Pranav and Sastry, Girish and Askell, Amanda and Agarwal, Sandhini and Herbert-Voss, Ariel and Krueger, Gretchen and Henighan, Tom and Child, Rewon and Ramesh, Aditya and Ziegler, Daniel and Wu, Jeffrey and Winter, Clemens and Hesse, Chris and Chen, Mark and Sigler, Eric and Litwin, Mateusz and Gray, Scott and Chess, Benjamin and Clark, Jack and Berner, Christopher and McCandlish, Sam and Radford, Alec and Sutskever, Ilya and Amodei, Dario},
 booktitle = {Advances in Neural Information Processing Systems},
 editor = {H. Larochelle and M. Ranzato and R. Hadsell and M.F. Balcan and H. Lin},
 pages = {1877--1901},
 publisher = {Curran Associates, Inc.},
 title = {Language Models are Few-Shot Learners},
 url = {https://proceedings.neurips.cc/paper_files/paper/2020/file/1457c0d6bfcb4967418bfb8ac142f64a-Paper.pdf},
 volume = {33},
 year = {2020}
}

@inproceedings{wu-etal-2024-reasoning,
    title = "Reasoning or Reciting? Exploring the Capabilities and Limitations of Language Models Through Counterfactual Tasks",
    author = {Wu, Zhaofeng  and
      Qiu, Linlu  and
      Ross, Alexis  and
      Akyurek, Ekin  and
      Chen, Boyuan  and
      Wang, Bailin  and
      Kim, Najoung  and
      Andreas, Jacob  and
      Kim, Yoon},
    editor = "Duh, Kevin  and
      Gomez, Helena  and
      Bethard, Steven",
    booktitle = "Proceedings of the 2024 Conference of the North American Chapter of the Association for Computational Linguistics: Human Language Technologies (Volume 1: Long Papers)",
    month = jun,
    year = "2024",
    address = "Mexico City, Mexico",
    publisher = "Association for Computational Linguistics",
    url = "https://aclanthology.org/2024.naacl-long.102",
    doi = "10.18653/v1/2024.naacl-long.102",
    pages = "1819--1862",
}

@inproceedings{
wei2022finetuned,
title={Finetuned Language Models are Zero-Shot Learners},
author={Jason Wei and Maarten Bosma and Vincent Zhao and Kelvin Guu and Adams Wei Yu and Brian Lester and Nan Du and Andrew M. Dai and Quoc V Le},
booktitle={International Conference on Learning Representations},
year={2022},
url={https://openreview.net/forum?id=gEZrGCozdqR}
}

@misc{yao2025preferenceleftbehindgroup,
      title={No Preference Left Behind: Group Distributional Preference Optimization}, 
      author={Binwei Yao and Zefan Cai and Yun-Shiuan Chuang and Shanglin Yang and Ming Jiang and Diyi Yang and Junjie Hu},
      year={2025},
      eprint={2412.20299},
      archivePrefix={arXiv},
      primaryClass={cs.CL},
      url={https://arxiv.org/abs/2412.20299}, 
}

@inproceedings{NEURIPS2022_b1efde53,
 author = {Ouyang, Long and Wu, Jeffrey and Jiang, Xu and Almeida, Diogo and Wainwright, Carroll and Mishkin, Pamela and Zhang, Chong and Agarwal, Sandhini and Slama, Katarina and Ray, Alex and Schulman, John and Hilton, Jacob and Kelton, Fraser and Miller, Luke and Simens, Maddie and Askell, Amanda and Welinder, Peter and Christiano, Paul F and Leike, Jan and Lowe, Ryan},
 booktitle = {Advances in Neural Information Processing Systems},
 editor = {S. Koyejo and S. Mohamed and A. Agarwal and D. Belgrave and K. Cho and A. Oh},
 pages = {27730--27744},
 publisher = {Curran Associates, Inc.},
 title = {Training language models to follow instructions with human feedback},
 url = {https://proceedings.neurips.cc/paper_files/paper/2022/file/b1efde53be364a73914f58805a001731-Paper-Conference.pdf},
 volume = {35},
 year = {2022}
}

@article{JMLR:v25:23-1042,
  author  = {Ruiqi Zhang and Spencer Frei and Peter L. Bartlett},
  title   = {Trained Transformers Learn Linear Models In-Context},
  journal = {Journal of Machine Learning Research},
  year    = {2024},
  volume  = {25},
  number  = {49},
  pages   = {1--55},
  url     = {http://jmlr.org/papers/v25/23-1042.html}
}

@inproceedings{dai-etal-2023-gpt,
    title = "Why Can {GPT} Learn In-Context? Language Models Secretly Perform Gradient Descent as Meta-Optimizers",
    author = "Dai, Damai  and
      Sun, Yutao  and
      Dong, Li  and
      Hao, Yaru  and
      Ma, Shuming  and
      Sui, Zhifang  and
      Wei, Furu",
    editor = "Rogers, Anna  and
      Boyd-Graber, Jordan  and
      Okazaki, Naoaki",
    booktitle = "Findings of the Association for Computational Linguistics: ACL 2023",
    month = jul,
    year = "2023",
    address = "Toronto, Canada",
    publisher = "Association for Computational Linguistics",
    url = "https://aclanthology.org/2023.findings-acl.247",
    doi = "10.18653/v1/2023.findings-acl.247",
    pages = "4005--4019",
}

@inproceedings{10.5555/3618408.3619217,
author = {Li, Yingcong and Ildiz, M. Emrullah and Papailiopoulos, Dimitris and Oymak, Samet},
title = {Transformers as algorithms: generalization and stability in in-context learning},
year = {2023},
publisher = {JMLR.org},
booktitle = {Proceedings of the 40th International Conference on Machine Learning},
articleno = {809},
numpages = {30},
location = {Honolulu, Hawaii, USA},
series = {ICML'23}
}

@misc{akyürek2023learningalgorithmincontextlearning,
      title={What learning algorithm is in-context learning? Investigations with linear models}, 
      author={Ekin Akyürek and Dale Schuurmans and Jacob Andreas and Tengyu Ma and Denny Zhou},
      year={2023},
      eprint={2211.15661},
      archivePrefix={arXiv},
      primaryClass={cs.LG},
      url={https://arxiv.org/abs/2211.15661}, 
}

@misc{
wei2024larger,
title={Larger language models do in-context learning differently},
author={Jerry Wei and Jason Wei and Yi Tay and Dustin Tran and Albert Webson and Yifeng Lu and Xinyun Chen and Hanxiao Liu and Da Huang and Denny Zhou and Tengyu Ma},
year={2024},
url={https://openreview.net/forum?id=DRGnEkbiQZ}
}

@article{10.1007/s00146-021-01327-5,
author = {Hoffmann, Christian Hugo},
title = {A philosophical view on singularity and strong AI},
year = {2022},
issue_date = {Aug 2023},
publisher = {Springer-Verlag},
address = {Berlin, Heidelberg},
volume = {38},
number = {4},
issn = {0951-5666},
url = {https://doi.org/10.1007/s00146-021-01327-5},
doi = {10.1007/s00146-021-01327-5},
journal = {AI Soc.},
month = jan,
pages = {1697–1714},
numpages = {18}
}

@inproceedings{10.1145/3442188.3445922,
author = {Bender, Emily M. and Gebru, Timnit and McMillan-Major, Angelina and Shmitchell, Shmargaret},
title = {On the Dangers of Stochastic Parrots: Can Language Models Be Too Big?},
year = {2021},
isbn = {9781450383097},
publisher = {Association for Computing Machinery},
address = {New York, NY, USA},
url = {https://doi.org/10.1145/3442188.3445922},
doi = {10.1145/3442188.3445922},
booktitle = {Proceedings of the 2021 ACM Conference on Fairness, Accountability, and Transparency},
pages = {610–623},
numpages = {14},
location = {Virtual Event, Canada},
series = {FAccT '21}
}

@inproceedings{bender-koller-2020-climbing,
    title = "Climbing towards {NLU}: {On} Meaning, Form, and Understanding in the Age of Data",
    author = "Bender, Emily M.  and
      Koller, Alexander",
    editor = "Jurafsky, Dan  and
      Chai, Joyce  and
      Schluter, Natalie  and
      Tetreault, Joel",
    booktitle = "Proceedings of the 58th Annual Meeting of the Association for Computational Linguistics",
    month = jul,
    year = "2020",
    address = "Online",
    publisher = "Association for Computational Linguistics",
    url = "https://aclanthology.org/2020.acl-main.463/",
    doi = "10.18653/v1/2020.acl-main.463",
    pages = "5185--5198"
}

@article{
doi:10.1073/pnas.2215907120,
author = {Melanie Mitchell  and David C. Krakauer },
title = {The debate over understanding in AI’s large language models},
journal = {Proceedings of the National Academy of Sciences},
volume = {120},
number = {13},
pages = {e2215907120},
year = {2023},
doi = {10.1073/pnas.2215907120},
URL = {https://www.pnas.org/doi/abs/10.1073/pnas.2215907120},
eprint = {https://www.pnas.org/doi/pdf/10.1073/pnas.2215907120}}

@misc{bubeck2023sparksartificialgeneralintelligence,
      title={Sparks of Artificial General Intelligence: Early experiments with GPT-4}, 
      author={Sébastien Bubeck and Varun Chandrasekaran and Ronen Eldan and Johannes Gehrke and Eric Horvitz and Ece Kamar and Peter Lee and Yin Tat Lee and Yuanzhi Li and Scott Lundberg and Harsha Nori and Hamid Palangi and Marco Tulio Ribeiro and Yi Zhang},
      year={2023},
      eprint={2303.12712},
      archivePrefix={arXiv},
      primaryClass={cs.CL},
      url={https://arxiv.org/abs/2303.12712}, 
}

@inproceedings{ganguli2022predictability,
  title={Predictability and surprise in large generative models},
  author={Ganguli, Deep and Hernandez, Danny and Lovitt, Liane and Askell, Amanda and Bai, Yuntao and Chen, Anna and Conerly, Tom and Dassarma, Nova and Drain, Dawn and Elhage, Nelson and others},
  booktitle={Proceedings of the 2022 ACM Conference on Fairness, Accountability, and Transparency},
  pages={1747--1764},
  year={2022}
}

@article{long2024taking,
  title={Taking AI welfare seriously},
  author={Long, Robert and Sebo, Jeff and Butlin, Patrick and Finlinson, Kathleen and Fish, Kyle and Harding, Jacqueline and Pfau, Jacob and Sims, Toni and Birch, Jonathan and Chalmers, David},
  journal={arXiv preprint arXiv:2411.00986},
  year={2024}
}

@misc{olsson2022incontextlearninginductionheads,
      title={In-context Learning and Induction Heads}, 
      author={Catherine Olsson and Nelson Elhage and Neel Nanda and Nicholas Joseph and Nova DasSarma and Tom Henighan and Ben Mann and Amanda Askell and Yuntao Bai and Anna Chen and Tom Conerly and Dawn Drain and Deep Ganguli and Zac Hatfield-Dodds and Danny Hernandez and Scott Johnston and Andy Jones and Jackson Kernion and Liane Lovitt and Kamal Ndousse and Dario Amodei and Tom Brown and Jack Clark and Jared Kaplan and Sam McCandlish and Chris Olah},
      year={2022},
      eprint={2209.11895},
      archivePrefix={arXiv},
      primaryClass={cs.LG},
      url={https://arxiv.org/abs/2209.11895}, 
}

@article{baron1985does,
  title={Does the autistic child have a “theory of mind”?},
  author={Baron-Cohen, Simon and Leslie, Alan M and Frith, Uta},
  journal={Cognition},
  volume={21},
  number={1},
  pages={37--46},
  year={1985},
  publisher={Elsevier}
}

@article{weizenbaum1966eliza,
  title={ELIZA—a computer program for the study of natural language communication between man and machine},
  author={Weizenbaum, Joseph},
  journal={Communications of the ACM},
  volume={9},
  number={1},
  pages={36--45},
  year={1966},
  publisher={ACM New York, NY, USA}
}

@misc{bonial2025dancingdeerconstructionalperspective,
      title={Dancing with Deer: A Constructional Perspective on MWEs in the Era of LLMs}, 
      author={Claire Bonial and Julia Bonn and Harish Tayyar Madabushi},
      year={2025},
      eprint={2508.15977},
      archivePrefix={arXiv},
      primaryClass={cs.CL},
      url={https://arxiv.org/abs/2508.15977}, 
}

@inproceedings{eisenschlos-etal-2023-winodict,
    title = "{W}ino{D}ict: Probing language models for in-context word acquisition",
    author = "Eisenschlos, Julian Martin  and
      Cole, Jeremy R.  and
      Liu, Fangyu  and
      Cohen, William W.",
    editor = "Vlachos, Andreas  and
      Augenstein, Isabelle",
    booktitle = "Proceedings of the 17th Conference of the European Chapter of the Association for Computational Linguistics",
    month = may,
    year = "2023",
    address = "Dubrovnik, Croatia",
    publisher = "Association for Computational Linguistics",
    url = "https://aclanthology.org/2023.eacl-main.7/",
    doi = "10.18653/v1/2023.eacl-main.7",
    pages = "94--102"
}

@article{levesque2012winograd,
  title={The Winograd schema challenge.},
  author={Levesque, Hector J and Davis, Ernest and Morgenstern, Leora},
  journal={KR},
  volume={2012},
  number={13th},
  pages={3},
  year={2012}
}
\bibliographystyle{icml2026}

\newpage
\appendix
\onecolumn
\section{Defining Reasoning: Bloom's Taxonomy}

We embrace a working definition of \textit{reasoning} based on the cognitive processes outlined in \citet{krathwohl2002revision}, who revises Bloom's original taxonomy of educational objectives \cite{bloom1956taxonomy}. Although the taxonomy's original purpose was to provide a framework for testing educational objectives, it is also a well-recognised rubric for defining and categorising levels of mastery within the cognitive domain generally, and has been leveraged within AI as well (e.g., \citet{blaha2022understanding}). Like the original taxonomy, Krathwohl's revised taxonomy breaks up the categories across the dimensions of \textit{Cognitive Process} and \textit{Knowledge}. 

In this research, we focus on the Cognitive Process dimension, which is an ordered set (from simple to more complex) of processes that subdivide reasoning: \textit{Remember, Understand, Apply, Analyse, Evaluate, Create}. 
Within these processes, we focus on \textit{Remember}, \textit{Understand} and \textit{Apply}. Our focus is guided by the fact that this is considered an ordered set of processes, such that the higher-level processes of \textit{Evaluate} and \textit{Create} rely upon mastery of the lower-level cognitive processes; within this classification we find evidence that LLMs are limited to \textit{Understand}, which would preclude higher-order processes.
\textit{Remember} is defined as ``Retrieving relevant knowledge from long-term memory,'' and is split into subcategories of \textit{Recognizing} and \textit{Recalling} \citet[p.~215]{krathwohl2002revision}. \textit{Understand} is defined as ``Determining the meaning of instructional messages, including oral, written, and graphic communication,'' and is split into finer-grained categories of \textit{Interpreting, Exemplifying, Classifying, Summarising, Inferring, Comparing,} and \textit{Explaining} (ibid.). \textit{Apply} is defined as ``Carrying out or using a procedure in a given situation,'' but we also leverage Bloom's original \textit{Application} description, which involves solving a new, unseen problem by deploying and operationalising previously learned information.  \textit{Apply} is split into \textit{Executing} and \textit{Implementing}.  

Because we argue that LLM reasoning is limited to \textit{Understand} and does not reach the level of \textit{Apply}, we will further explain the distinction between these two processes. If we assume a set of items that have been labelled as being members of a certain class, then \textit{Understanding} this information assumes that one can explain why the items are members of that class, as well as compare the known set of items to other items in order to further classify items and provide examples of the class. This is similar to teaching a student about the parts of speech of their language, then asking the student to explain what a verb is, provide some examples of verbs, and recognise whether or not a familiar word is a verb. To engage in the \textit{Apply} process, one must first \textit{Understand}, but then also determine how to deploy that understanding \emph{with respect to a new set of items where the generalisation that makes this set a class is not already given}.  This is similar to asking a linguist to examine the morpho-syntax of a new language and determine what the parts of speech are in that language---the linguist will generalise knowledge of the features shared by members of part-of-speech classes in other languages and apply this knowledge to the new language.  

Finally, \textit{Analyse} is defined as ``Breaking material into its constituent parts and detecting how the parts relate to one another and to an overall structure or purpose,'' and is split into \textit{Differentiating, Organising} and \textit{Attributing}.  

The Knowledge dimension is ordered with respect to more concrete to more abstract types: \textit{Factual Knowledge, Conceptual Knowledge, Procedural Knowledge, Metacognitive Knowledge}.  In the present research, we focus on \textit{Factual Knowledge} and \textit{Conceptual Knowledge}.  \textit{Factual Knowledge} is defined as ``The basic elements that students must know to be acquainted with a discipline or solve problems within it'' and is split into \textit{Knowledge of terminology} and \textit{Knowledge of specific details and elements} \citet[p.~215]{krathwohl2002revision}.  \textit{Conceptual Knowledge} is defined as ``The interrelationships among the basic elements within a larger structure that enable them to function together'' and includes subcategories of \textit{Knowledge of classification and categories, Knowledge of principles and generalisations,} and \textit{Knowledge of theories, models, and structures} (ibid.). 

In terms of this theoretical framework, we argue that LLM reliance on pre-training data enables \textit{Remembering} of \textit{Factual Knowledge}, which necessarily requires memorisation as opposed to a deeper level of the Cognitive Process dimension. Furthermore, LLMs certainly \textit{Remember} an immense amount of \textit{Conceptual Knowledge}, including the memorisation of theories and even potentially commonplace generalisations; we emphasise that such \textit{Remembering} may, to the human eye, masquerade as \textit{Understanding} and even \textit{Applying}, as people are more likely to apply a heuristic in lieu of memorising web-scale information. Thus, we may attribute similar cognitive processes to LLMs where they are in fact able to rely on memory.

Nonetheless, in effectively \textbf{extrapolating}
from immense amounts of pre-training data, LLMs are able to demonstrate \textit{Understanding} of knowledge that is well-represented in pre-training data. Extrapolation from pre-training data is a combination of \textit{Remembering}, \textit{Understanding} and hallucination.  We argue that when the pre-training data closely resembles the target task, memory is sufficient for what is essentially stochastic parroting. When the prompt includes contextual direction, either through ICL examples in base models, or through the prompt itself in instruction-tuned models, then models are able to effectively extrapolate from their pre-training data, demonstrating some \textit{Understanding} as defined above. Finally, if the pre-training data to be extrapolated from is not effectively triggered by the prompt (either examples or directly through the prompt) or that data simply is not present, models will hallucinate. While not factually correct, hallucinations may still maintain the illusion of deeper cognitive processes, especially to the human eye, which is likely to attribute human-like reasoning (the `ELIZA effect' \cite{weizenbaum1966eliza}). 

Overall, we argue and provide evidence that LLM reasoning is limited to what is termed \textit{Understanding} in Bloom's framework. 
We also argue that we \emph{cannot} test whether or not the reasoning process of \textit{Apply} takes place unless i) we can guarantee that \textit{Remembering} is not the likely explanation given the relevance of pre-training data to answering examples in the test set, and ii) that successful performance on evaluations cannot be attributed to contextual-guided extrapolation, which simplifies the task to the basic \textit{Understanding} processes, including \textit{Classifying} (the inverse of \textit{Exemplifying}), as well as \textit{Comparing} test items to the provided examples.

\subsection{Terminological Notes}
We caution the reader against assuming commonplace definitions of these terms---there are certainly levels of ``understanding'' that models do not achieve, and it is clear that models can do some types of evaluation and can create new texts, but this does not necessarily meet the definitional criteria of these terms as used in educational settings.

We note that the term extrapolation is also sometimes applied to human cognitive processes; however, as described in that literature, it is a more advanced reasoning process than mathematical extrapolation. For example, \citet{aviles2000teaching} describes extrapolation as a part of \textit{Understanding} (termed `Comprehension' in Bloom's original taxonomy), and indicates that it involves identifying trends and consequences (e.g., describing the intended and unintended consequences of social policies). Within AI, identifying consequences is considered more advanced causal reasoning, and, in the case of a potential social policy, this would be akin to counterfactual reasoning, as opposed to extrapolation \cite{pearl2000models}.

The phrase stochastic parroting does not have a precise parallel in the literature on human cognitive processes described in Section~\ref{sec:framework}. It relies strongly on \textit{Remembering}, but rather than memory of chunks of information, it also involves extrapolation to guess the next most likely token, based upon what is in memory.  In the present research, the term extrapolation refers to the mathematical, machine-learning sense: estimating values outside the range of known data points based on the observed relationship between variables.

\section{Context-Directed Extrapolation vs Stochastic Parroting}
\label{app:cdevssp}
We would like to acknowledge that Context-Directed Extrapolation and Stochastic Parroting share some commonalities. Therefore, this section clarifies the commonalities and differences between the stochastic parroting perspective and the perspective of context-directed extrapolation from training data priors.

\textbf{Commonalities.} From the perspective of base model performance, there is no functional difference between context-directed extrapolation and stochastic parroting. As previously discussed, base models consistently fail on tasks requiring reasoning when they are presented without examples. One can argue that the examples simply form a long context, making the correct answer the most probable sequence completion. This makes both theories appear to describe the same mechanism: the model completes a given sequence based on statistical patterns. \emph{Consequently, when analysing base models alone, the two views are indistinguishable.}

\textbf{Differences.} The functional divergence between these two views becomes apparent with instruction-tuned models. While a base model's output can be seen as the most probable completion of a long sequence of examples, the context-directed extrapolation framework posits that instruction-tuning enables a different mechanism. Our framework suggests that instruction-tuning allows a model to interpret an instruction as a directive to construct an \emph{implicit} task context, rather than as a literal sequence to be continued. This context activates relevant priors from the model's pre-training data, enabling it to solve the specified task. The critical evidence supporting this distinction is that instruction-tuned models can solve logical deduction problems without any examples. This phenomenon is directly accounted for by context-directed extrapolation but is not straightforwardly explained without additional assumptions by a strict stochastic parroting model. This functional gap becomes even more stark in tasks involving novel words, as described previously.

\textbf{Grounding} An important implication of context-directed extrapolation is that it facilitates a limited form of grounding. This is not grounding in the sense of connecting language to embodied experience, but rather a mechanism by which the model can access information that is not explicitly present in the surface text. Because the process involves extrapolating from priors activated by the prompt, a model can, for instance, take the definition of a nonce verb and productively apply that new meaning in novel contexts.

Therefore, while there is some overlap, our context-directed perspective is fundamentally distinct from a stochastic parroting mechanism, which does not naturally predict reliable behaviour on novel inputs that fall outside its memorised distribution. The demonstrated ability of LLMs to extend prompt-based definitions, apply abstract patterns, and generate context-appropriate interpretations indicates that extrapolation provides access to information not reducible to surface statistics. In this sense, context-directed extrapolation offers a framework for explaining LLM capabilities and limitations, including a form of limited grounding, which remain contingent upon the model's training priors and the context supplied at inference.

\end{document}